\documentclass[conference]{IEEEtran}
\IEEEoverridecommandlockouts
\usepackage{cite}
\usepackage{amsmath,amssymb,amsfonts}
    
\usepackage[pdftex]{graphicx}
\usepackage{subcaption}
\usepackage{hyperref}
\usepackage{url}
\usepackage{algorithm}
\usepackage[noend]{algpseudocode}
\usepackage{color}

\usepackage{zref-base}
\usepackage{atveryend}
\makeatletter
\AfterLastShipout{%
  \if@filesw   
    \begingroup
      \zref@setcurrent{default}{\the\value{equation}}%
      \zref@wrapper@immediate{%
        \zref@labelbyprops{eq-last}{default}%
      }%
    \endgroup
  \fi
}
\makeatother

\usepackage{xr}
\makeatletter
\def\BState{\State\hskip-\ALG@thistlm}
\makeatother

\newtheorem{conjecture}{Conjecture}
\newcommand*{\QEDB}{\hfill\ensuremath{\square}}%

\begin{document}

\title{Entropy-Constrained Training of\\ Deep Neural Networks}

\author{\IEEEauthorblockN{1\textsuperscript{st} Simon Wiedemann}
\IEEEauthorblockA{\textit{Dept. of Video Coding and Analytics} \\
\textit{Fraunhofer HHI}\\
Berlin, Germany \\
simon.wiedemann@hhi.fraunhofer.de}
\and
\IEEEauthorblockN{2\textsuperscript{nd} Arturo Marban}
\IEEEauthorblockA{\textit{Dept. of Video Coding and Analytics} \\
\textit{Fraunhofer HHI}\\
Berlin, Germany \\
arturo.marban@tu-berlin.de}
\and
\IEEEauthorblockN{3\textsuperscript{rd} Klaus-Robert M\"uller}
\IEEEauthorblockA{\textit{Machine Learning Group} \\
\textit{TU Berlin}\\
Berlin, Germany \\
klaus-robert.mueller@tu-berlin.de}

\and
\IEEEauthorblockN{4\textsuperscript{th} Wojciech Samek}
\IEEEauthorblockA{\textit{Dept. of Video Coding and Analytics} \\
\textit{Fraunhofer HHI}\\
Berlin, Germany \\
wojciech.samek@hhi.fraunhofer.de}

}

\maketitle

\begin{abstract}
We propose a general framework for neural network compression that is motivated by the Minimum Description Length (MDL) principle. For that we first derive an expression for the entropy of a neural network, which measures its complexity explicitly in terms of its bit-size. Then, we formalize the problem of neural network compression as an entropy-constrained optimization objective. This objective generalizes many of the compression techniques proposed in the literature, in that pruning or reducing the cardinality of the weight elements of the network can be seen special cases of entropy-minimization techniques. Furthermore, we derive a continuous relaxation of the objective, which allows us to minimize it using gradient based optimization techniques. Finally, we show that we can reach state-of-the-art compression results on different network architectures and data sets, e.g. achieving x71 compression gains on a VGG-like architecture. 
\end{abstract}

\section{Introduction}
It is well established that deep neural networks excel on a wide range of machine learning tasks \cite{DeepLearning}. However, common training practices require to equip neural networks with millions of parameters in order to attain good generalisation performances. This often results in great storage requirements for representing the networks (in the order of hundreds of MB), which greatly difficults (or even prohibits) their deployment into resource constrained devices and their transmission in restricted communication channels. Moreover, several recent works \cite{DLC_survey} have shown that most deep networks are overparametrized for the given task and that they have the capacity to memorize entire data sets \cite{Random_labels}. This motivates the use of regularizers that penalize the networks complexity, lowering the risk of overfitting and thus, favour models that generalize well.

Compression, that is, lowering the amount of information required to uniquely reconstruct the network's parameters, is a very natural way to limit the network's complexity. Indeed, the Minimum Description Length (MDL) principle states that if two or more models achieve same prediction performance, then one should always choose the model that requires the least amount of bits to describe it \cite{Rissanen_MDL, MDL}. Moreover, compression also offers direct practical advantages. Networks with small memory footprint do not only require lower communication costs (which can be of utmost importance in, e.g., federated learning scenarios \cite{federated_learning, SatArXiv18}), but can also perform inference requiring significantly less time and energy resources \cite{DLC_survey}.

Furthermore, we know from the source coding literature \cite{Shannon, Cover, Wiegand_source_coding} that the {\it entropy} of a probabilistic source measures the minimum average bit-length needed in order to compress the data generated by that source in a lossless manner. In addition, recent work have focused on designing efficient data structures that exploit the fact that the network has low entropy\footnote{The entropy is measured relative to the empirical probability mass distribution of the weight elements.}, achieving lower memory, time and energy footprint for performing inference \cite{deep_compression, universal_succint_dnn_compression, Universal_dnn_compression, Simon_lossless_dnn_compression1}. Hence, due to theoretical as well as practical reasons, it appears only natural to measure the bit-size of neural networks in terms of their entropy and to constrain their training explicitly by it.

Hence, our contributions can be summarized as follows:
\begin{itemize}
\item We formulate the task of neural network compression under an entropy-constrained optimization objective. The derived entropy term explicitly measures the minimum average bit-size required for representing the weight elements, generalizing many of the compression techniques proposed so far in the literature.
\item We derive a continuous relaxation of the entropy-constrained objective, which allows to minimize it using gradient-based optimization techniques. We conjecture that the continuous objective upper bounds the desired discrete one, and provide experimental evidence for it.   
\item Our experimental results show that we can highly compress different types of neural network architectures while minimally impacting their accuracy on different data sets.
\end{itemize}

\section{The MDL principle as a general framework for deep neural network compression}
\label{sec: MDL principle for DNN}
We start by formalising the general learning problem under the MDL principle, since the objective of model compression arises naturally under this paradigm. 

Assume a supervised learning setting where, given a particular data set $\mathcal{D}_N = \{(x_i, y_i) | x_i \in \mathcal{X}, y_i \in \mathcal{Y}, i\in \{1,..., N=|\mathcal{X}|=|\mathcal{Y}|\}\}$, we want to learn to predict the elements $y_i$ of the output (or label) set $\mathcal{Y}$ from the elements $x_i$ of the input set $\mathcal{X}$. The MDL principle equates the task of ``learning'' with the task of compression. That is, the more regularities we find in the data, the more we can use these to compress it and ultimately, generalize better to unseen data that entail the same regularities. Therefore, the above supervised learning goal can be restated as to find the compression algorithm (or code) that compresses the most of the set of labels $\mathcal{Y}$ in a lossless manner given the input set $\mathcal{X}$. 

Due to the fundamental correspondence between probability distributions and bit-lengths \cite{MDL}, we can precisely formalize this idea. Namely, let $\mathcal{W} = \{p(\mathcal{Y}|\mathcal{X}, W)| W\in \mathbb{R}^n, n\in \mathbb{N}\}$ be a set of parametric conditional probability distributions over the label set $\mathcal{Y}$, where we denote as $W \equiv p(\mathcal{Y}|\mathcal{X}, W)$ a particular model or point hypothesis in $\mathcal{W}$ (e.g., a trained neural network). Then, one possible way to encode the labels $\mathcal{Y}$ is by using a so called \textit{two-part code}. That is, we first describe a particular point hypothesis $W$ in $\mathcal{W}$ using $L_{\mathcal{W}}(W)$ bits and then optimally encode its prediction errors using $L(\mathcal{Y}|\mathcal{X}, W)=-\log_2{p(\mathcal{Y}|\mathcal{X}, W)}$ bits. Hence, the MDL principle states that we should find
\begin{equation}
W^* = \min_W \underbrace{-\log_2 p(\mathcal{Y}|\mathcal{X}, W)}_{encoding\ prediction\ error} +\ \alpha \underbrace{L_{\mathcal{W}}(W)}_{encoding\ model}
\label{Eq: crude MDL}
\end{equation}
where $0 < \alpha \in \mathbb{R}$.

Notice, that \eqref{Eq: crude MDL} aims to minimize the prediction errors of the model \textit{and} the explicit bit-length of the model. Therefore, the MDL principle is a natural framework for the task of neural network compression.
Now we need to define a description or code that assigns an unique bit-string to each model $W$.

Firstly, we consider the entire set of discrete neural networks $\mathbb{W}$ with a particular and fixed topology. That is, all considered neural networks have same topology which does not change over time, and their weight element values are restricted to be elements from a finite set of real numbers. Thus, $w_i$ denotes the $i$-th weight element of the neural network, which can take one of $K$ possible values, i.e., $\omega_i \in \Omega = \{\omega_0,..., \omega_{K-1}\}$. We furthermore assume that the network has $n$ parameters, i.e., $i \in \{1,...,n\}$. Whilst this definition may seem a considerable constrain  at first sight, notice that $\mathbb{W}$ entails all neural networks that are representable by any digital device\footnote{Assuming a sufficiently large $K$.}. Moreover, focusing on networks with fixed topology is reasonable because: 1) storing the particular topology should be negligible compared to storing the weight values and 2) in many practical settings one aims to compress a network parameters with an already given topology (e.g., communication scenario).

Secondly, we assume that the decoder (or decompression algorithm) has no concrete knowledge regarding the mutual information between the input and output set. Consequently, we cannot make any reasonable a priori assumptions regarding the correlation between the weight values of the neural network. This scenario is very common in most real world cases. For instance, most digital processor units implicitly make such assumption by not prioritising any particular network configuration. Moreover, this is also an usual assumption in machine learning problems.

Consequently, a reasonable prior over the weight elements is the $n$-long discrete independent random process. That is, let $P_{\Omega} = \{p_0, ..., p_{K-1}\}$ be a particular probability mass distribution (PMD) over the finite set $\Omega$. Then, we model the joint probability distribution as $P(W) = P(w_1=\omega_{k_1},..., w_n=\omega_{k_n}) = \prod_i^n p_{k_i}$, where $p_{k_i} \in P_{\Omega}$ is the probability of the $i$-th weight element outputting the value $\omega_{k_i} \in \Omega$. Hence, the respective bit-length can be defined as $L_{\mathbb{W}}(W) = -\log_2P(W) = \sum_i^n-\log_2{p_{k_i}}$.  However, the resulting code depends on the choice of the PMD, which we don't know a priori. 

Thus, a possible choice for PMD is to assume an uniform prior over $\Omega$, i.e., $p_k = 1/K$. However, it assigns a bit-string of constant size to each weight element and consequently to each model $W$. Concretely, $L_{\mathbb{W}}(W) = n\log_2K$, and if $K = 2^{32}$ we obtain the amount of bits needed to store the network if we store each element in, e.g., its respective single precision floating point representations. In this case the minimization \eqref{Eq: crude MDL} becomes the standard maximum likelihood estimation objective and consequently no model compression is performed in the minimization process.

However, we can design more suitable codes in the case of deep neural networks. Namely, for each particular neural network $W$, we could first estimate the probability distributions of its weight values and subsequently use these estimates in the compression process. This may work well since the number of parameters in most state-of-the-art deep neural network models is typically $n=\mathcal{O}(10^7)$ and therefore, we can expect the estimation to be  a good approximation of the real distribution. Hence, we propose to first calculate the maximum likelihood estimate of the PMD of $W$, and then optimally encode the element values of $W$ relative to it. Such coding techniques belong to the subject of \textit{universal coding} and their properties are well studied in the literature \cite{Rissanen_MDL, MDL_coding, MDL_assymptotic_conv, MDL}. Concretely, from \cite{MDL} we know that if we apply an universal two-part code for the task probability density estimation by using histograms, the respective regret (or maximum redundancy) per data point decreases sublinearly as the number of  data points increases. That is, the maximum redundancy of the code decreases at a rate of order $\mathcal{O}(\frac{\log_2n}{n})$, $n$ being the number of data points. This means that we can expect the regret to be almost 0 if we apply such a code on to the weights of deep neural networks, meaning that the empirical PMD is almost identical to the actual PMD that generated the weight values (assuming that such a distribution actually generated the weight values). 

\subsection{Universal coding of discrete deep neural networks}
\begin{figure}[t]
\includegraphics[width=\columnwidth,clip,keepaspectratio]{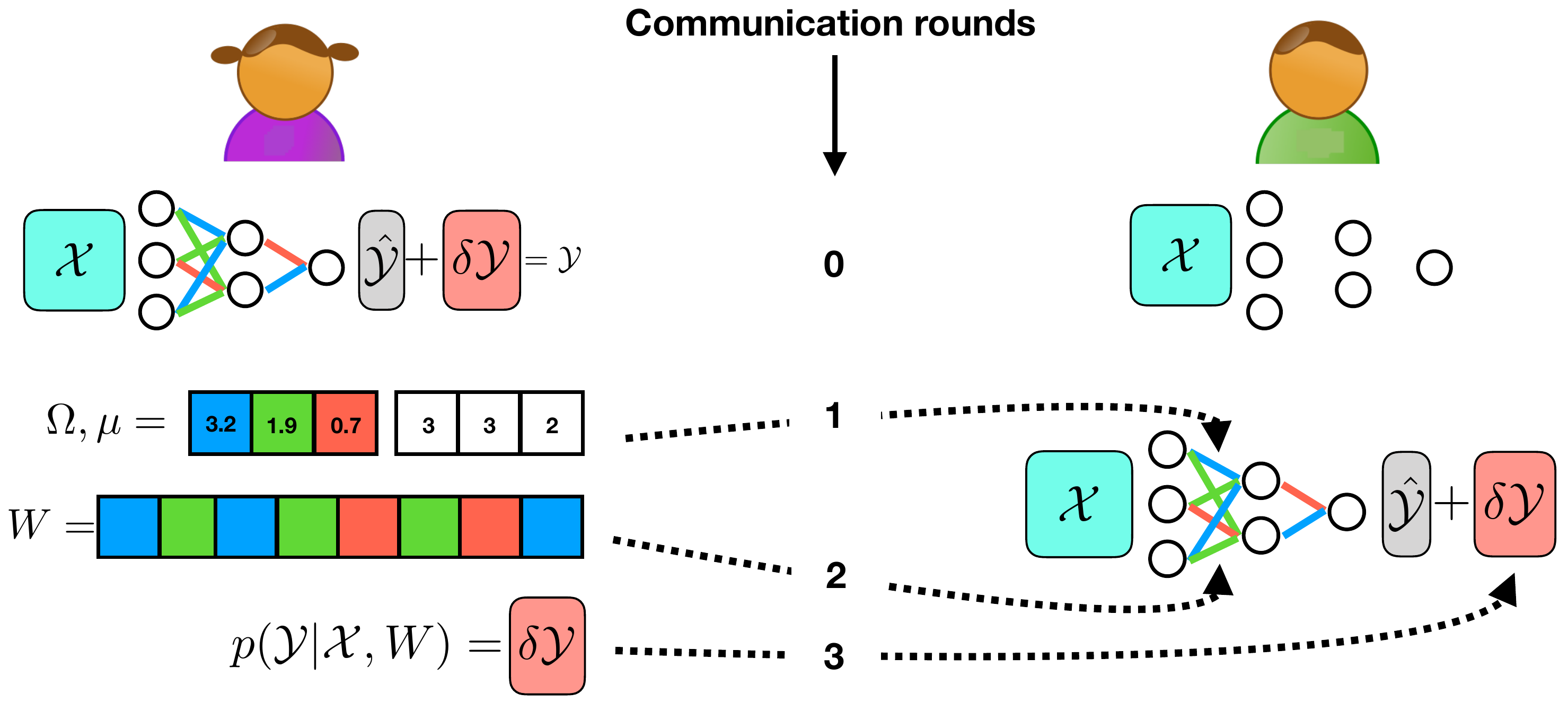} 
\caption{The task of compression is analogous to the task of communication, which is depicted in this figure. Alice wants to send to Bob a particular output set $\mathcal{Y}$. Assume that Alice and Bob have previously agreed on the input set $\mathcal{X}$ and a particular neural network topology. Then, Alice will send $\mathcal{Y}$ performing three steps: 1) She trains a particular network model $p(\mathcal{Y}|\mathcal{X}, W) \equiv W$, encodes the unique elements $\Omega = \{3.2,1.9,0.7\}$ (blue, green, red) that appear in the network using an uniform code, along with their maximum likelihood estimate of their probability mass distribution $\mu=(3,3,2)$, and sends these to Bob. 2) She encodes the weight values using an entropy-coder with regards to $\mu$ and sends them to Bob. Since Bob also knows $\mu$ and $\Omega$, he can uniquely recover the weight values. 3) Alice encodes the prediction errors of $W$ using $-\log_2p(\mathcal{Y}|\mathcal{X}, W)$ bits and sends these to Bob. With that, Bob can uniquely reconstruct the output set $\mathcal{Y}$.}
\label{Fig: NN compression}
\end{figure}

Hence, we propose the following coding scheme:  Let $\mu = \{\hat{p}_0, ..., \hat{p}_{K-1} \}$ denote the empirical probability mass distribution (EPMD) of the elements $\Omega$ in $W$, thus, $\hat{p}_k = \#(\omega_k)/n$ where $\#(\cdot)$ denotes the counting operator. Then, we 
\begin{enumerate}
\item calculate $\mu$ and encode its elements along with $\Omega$ using an uniform code, and
\item compress each $w_i$ using $-\log_2 \sum_k I(i=k)\mu_k$ bits respectively.
\end{enumerate}
Here, $I(i=k)$ denotes the indicator function, being 1 if $w_i = \omega_k$ and 0 else. Figure \ref{Fig: NN compression} sketches this coding scheme. 

Thus, the resulting bit-length of a particular neural network model $W$ can be written as
\begin{equation}
L_{\mathbb{W}}(W) = L(\mathbb{I}_W) + L(\mu) + L(\Omega)
\label{Eq: W bit length}
\end{equation}
where $ L(\mathbb{I}_W) = \sum_{i=1}^n -\log_2{\sum_k I(i=k)\mu_k}$, $L(\mu)=K\log_2{n}$ and $L(\Omega)=Kb$ with $b$ being the bit-precision used to represent the elements in $\Omega$.

\subsection{Entropy-constrained minimization objective}
Notice, that $ L(\mathbb{I}_W) = n\sum_k -\mu_k\log_2\mu_k = nH(\mu)$, with $H(\mu)$ being the entropy of the EPMD $\mu$. Thus, \eqref{Eq: W bit length} states that the bit-length of each weight element $w_i$ is equal to the minimum average bit-length relative to $\mu$, plus additional redundant terms that result from having to send the estimates.
Hence, since the bit-length of the estimation $L(\mu)$ and $L(\Omega)$ is constant, the MDL principle states that the optimally compressed neural network minimizes the objective
\begin{equation}
W^* = \min_{W\in \mathbb{W}} -\log_2p(\mathcal{Y}|\mathcal{X}, W) + \alpha nH(\mu), \quad 0 < \alpha \in \mathbb{R}
\label{Eq: DNN min objective}
\end{equation} 

As a side note we want to mention that most common lossless entropy coders encode the outputs of an independent random process in the same manner as described above \cite{Wiegand_source_coding}. Thus, the above description of the neural network explicitly expresses the amount of bits required to store its weight elements if we would apply, e.g. 
the Huffman code or an arithmetic coder for encoding its elements.

\section{Training of deep neural networks under an entropy-constrained objective}
We would now like to apply gradient-based optimization techniques in order to minimize \eqref{Eq: DNN min objective}, since they are scalable to large and deep networks and have shown to be able to find good local minima that generalize well \cite{DL_tricks_trade, DeepLearning}. However, the minimisation objective as stated in \eqref{Eq: DNN min objective} is discrete and non-differentiable. Therefore, in the following we firstly derive a continuous relaxation of \eqref{Eq: DNN min objective} and subsequently reformulate it as variational minimization objective. Then, we will conjecture that, under certain conditions, the variational objective upper bounds the discrete one.

\subsection{Continuous relaxation}
Notice that the only reason \eqref{Eq: DNN min objective} is non-differentiable is due to the indicator operator implicitly entailed in it. Hence, we propose to smooth it by a differentiable discrete probability distributions. Namely,
\begin{align}
 I(i=k) = & \begin{cases}
1, & \text{ if } w_i = \omega_k \\
0, & \text{ else}
\end{cases} \nonumber \\
\xrightarrow{\text{cont. relax.}} P_i = & \{P(i=k|\theta_{ik})|  \sum_k P(i=k|\theta_{ik}) = 1\}
\label{Eq: Cont relax of indicator}
\end{align}
with $P(i=k|\theta_{ik})=P_{ik}$ being the probability that the weight element $w_i$ takes the discrete value $\omega_k$, parametrized by $\theta_{ik}$. Thus, the continuous relaxation now models an entire set of discrete neural networks, whose probability of being sampled is specified by the joint probability distribution $P_{\theta} = \prod_i^n P_i$.

Subsequently, we naturally replace  each estimate $\mu_k$ of the PMD, and thus the entropy, by
\begin{align}
\mu_k = \frac{1}{n}\sum_i I(i=k) &\xrightarrow{\text{cont. relax.}}  P_k = \frac{1}{n}\sum_i P_{ik} \nonumber \\
H(\mu) = \sum_k -\mu_k\log_2\mu_k & \xrightarrow{\text{}}  H(P) = \sum_k -P_k\log_2P_k
\label{Eq: Cont relax entropy}
\end{align}
and reformulate the minimization objective \eqref{Eq: DNN min objective} as
\begin{equation}
\theta^* = \min_{\theta} \mathbb{E}_{P_{\theta}}[-\log_2p(\mathcal{Y}|\mathcal{X},W\sim  P_{\theta})] + \alpha nH(P)
\label{Eq: DNN min variational cont relax}
\end{equation}
Thus, we now aim to minimize the \textit{averaged} prediction error of the network, as taken relative to our model $P_{\theta}$, constrained by the respective relaxation of the entropy \eqref{Eq: Cont relax entropy}.

However, we can still not apply gradient-based optimization techniques in order to minimize \eqref{Eq: DNN min variational cont relax} as calculating the mean of the log-likelihood is infeasible for deep neural networks.

\subsection{Training neural networks under the entropy-constrained variational objective}
\begin{algorithm}[t]
\caption{Forward pass sampling from discrete probability distribution of the layers weight tensor}
\label{Alg: noise forward pass}
\begin{algorithmic}[1]
\Procedure{Forwardpass}{a} \Comment{per layer}
\State Calculate mean weight tensor $\nu_W$ and respective variance $\sigma_W^2$.
\State Forward pass layers mean $\nu_z$ and standard deviation $\sigma_z$ as specified in \eqref{Eq: mean weight} and \eqref{Eq: var weight}.
\State Sample a tensor $\epsilon \sim N(0,1)$ with the same dimension as the preactivation layers.
\State Calculate preactivation values as $Z = \nu_z + \sigma_z\odot \epsilon$ \Comment{$\odot $ denotes the Hadamard product}.
\EndProcedure
\end{algorithmic}
\end{algorithm}
Hence, in order to train the network under the objective \eqref{Eq: DNN min variational cont relax} we apply similar scalable approximation techniques as proposed by the bayesian neural networks literature \cite{variational_dropout, weight_info_dropout, VD_sparsifies, BayesianCompression}. Concretely, we approximate the mean of the log-likelihood by an unbiased Monte-Carlo estimator and sample from the preactivation values of the network, whose mean and variance depend on the mean and variances of the weight element and the activation values of the respective layer. That is, each preactivation value is now modeled as $Z = \nu_z + \sigma_zN(0,1)$, where
\begin{align}
\nu_z & = \nu_W\cdot a \label{Eq: mean weight}\\
\sigma_z & = \sqrt{\sigma_W^2\cdot a^2} \label{Eq: var weight}
\end{align}
and $(\nu_W)_{i} = \sum_k \omega_kP_{ik}$, $(\sigma_W)^2_{i} = \sum_k \omega_k^2P_{ik} -(\nu_W)_i$ are the mean and variances of the weights of the respective layer, and $a$ are the activation values of the previous layer.

From a source coding point of view this procedure simulates a stochastic quantization scheme of the weights relative to the joint probability model $P_{\theta}$, which can be expressed as Gaussian noise in the preactivation layers due to the central limit theorem. The respective pseudocode can be seen in algorithm \ref{Alg: noise forward pass}.
Notice, that now \eqref{Eq: DNN min variational cont relax} is differentiable with respect to the parameters $\theta$ and discrete values $\Omega$.

\begin{figure}[t]
\includegraphics[width=\columnwidth,clip,keepaspectratio]{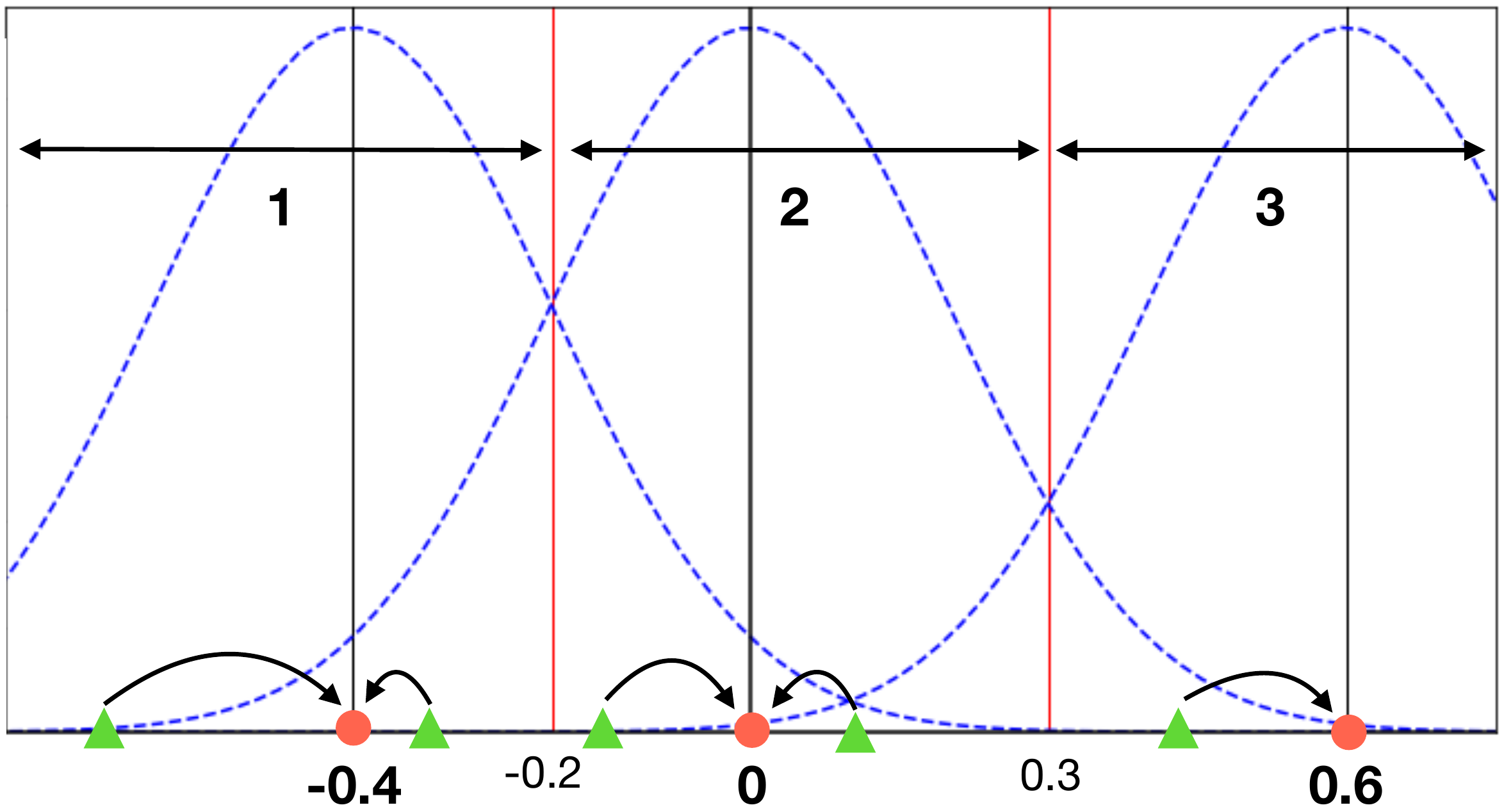} 
\caption{Quantization of the weight values corresponds to the maximum-a-posteriori (MAP) operator applied to their parametric discrete probability distribution. In our experiments we parametrized the probability distributions using Gaussian kernels, whose MAP operator is equivalent to a nearest-neighbor quantization scheme, as depicted in the above diagram. For instance, all continuous weight element values (green triangles) that lie in region 1 will be quantized to the maximum probable value of that region, namely the value $\omega_1 = -0.4$ (red dot). Analogously for the regions 2 and 3. For constant variance, the decision thresholds (red lines) lie exactly in the middle between two elements of $\Omega = \{-0.4, 0, 0.6\}$.}
\label{Fig: weight quantization}
\end{figure}

After training, we then select the most likely point hypothesis as our discrete model.  Henceforth, we will refer to this operation as the maximum-a-posteriori (MAP) quantization step, and denote it as $W^* = q(\theta^*)$. If all $P_{ik}$ are defined in terms of a radial function relative to a continuous parameter $\theta_i$ and $\omega_k$, then the MAP quantization corresponds to a nearest-neighbor quantization scheme of the parameter $\theta_i$ relative to the discrete set $\Omega$ (as depicted in figure \ref{Fig: weight quantization}).

\subsection{Relation between the variational and discrete objectives}
The motivation behind minimizing \eqref{Eq: DNN min variational cont relax} is not only due to the scalability property, but also because we conjecture that it upper bounds the loss value \eqref{Eq: DNN min objective} of the MAP point hypothesis $W^*$ for sufficiently small loss values.
 \vspace*{0.25cm}
 
\begin{conjecture}
Let $\theta^*$ be the parameters of the joint probability model $P_{\theta^*} = \prod_i^n P_i$ which minimizes the variational objective \eqref{Eq: DNN min variational cont relax}, where each $P_i$ is defined as in \eqref{Eq: Cont relax of indicator}. Then, if its variational loss value is sufficiently small, it upper bounds the loss value of the respective MAP point hypothesis $W^* = q(\theta^*)$. That is, there exist a $0 < \epsilon\in \mathbb{R}$, where 
 \begin{align}
 &  \; \mathcal{L}(P_{\theta^*}) = \mathbb{E}_{P_{\theta^*}}[-\log_2p(\mathcal{Y}|\mathcal{X}, W\sim  P^*_{\theta})] + \alpha nH(P(\theta^*)) \nonumber \\
 >& \;  \mathcal{L}(W^*) = -\log_2p(\mathcal{Y}|\mathcal{X}, W^*)  + \alpha nH(\mu^*) 
  \label{Eq: conj upper bound}
 \end{align}
with high probability, if $\mathcal{L}(P_{\theta^*}) < \epsilon$. $\mu^*$ refers to the EPMD of the point hypothesis $W^*$.
\label{Conj: upper bound}
\end{conjecture}

Although we do not provide a rigorous proof of conjecture  \ref{Conj: upper bound},  in the following we give a proof idea in order to support the claim stated in it.
\vspace*{0.25cm}

\textit{Proof idea:}
Firstly, notice that $P_{\theta} \xrightarrow{\text{var}(P_{\theta})\rightarrow 0} \mu$. That is, the PMD of the continuous relaxation $P_{\theta}$ converges to the EPMD $\mu$ of the MAP point hypothesis in the limit of zero variance. Consequently, we argue that the entropy of the EPMD $H(\mu)$ is always smaller or equal to the entropy of its continuous relaxation $H(P)$ (as defined in \eqref{Eq: Cont relax entropy})
\[H(P) \geq H(\mu)\]
The intuition behind the above inequality is that we can reduce the entropy $H(P)$ by reducing the variance of each element of the random process $P_{\theta}$ (thus, the variance of each weight element) and, in the limit where each element has 0 variance, we obtain the MAP distribution $\mu$.

Secondly, we argue that the mean value of the log-likelihood is with high probability larger than the prediction error of the MAP estimate at the minima. This becomes more clear if we rewrite the mean as
\begin{align}
 & \mathbb{E}_{P_{\theta}}[-\log_2p(\mathcal{Y}|\mathcal{X},W\sim  P_{\theta})] = L^* P_1 +\bar{L}(1-P_1)
\label{Eq: mean log likelihood}
\end{align}
where $ L^* = -\log_2p(\mathcal{Y}|\mathcal{X},W^*)$ is the prediction error of the MAP point estimate, $P_1$ its respective probability of being sampled and $\bar{L} = \mathbb{E}_{P_{\theta} \setminus P_1}[-\log_2p(\mathcal{Y}|\mathcal{X},W\sim  P_{\theta}\setminus P_1)]$  the mean of the prediction errors as taken relative to all the other point estimates and the respective conjugate probabilities $ P_{\theta}\setminus P_1$. Hence, the likelihood of the MAP point estimate is smaller iff $L^* < \bar{L}$. This is likely to be true iff the variance of $P_{\theta}$ is sufficiently small, since $P_1 \xrightarrow{\text{var}(P_{\theta})\rightarrow 0} 1$ \footnote{This argument can be relaxed into $P_1$ expressing the probability of the set of points $W_j$ near $W^*$, such that their respective loss $L_j \approx L^*$.}.  

Lastly, minimizing \eqref{Eq: conj upper bound} implies minimizing the entropy of $P_{\theta}$ which implicitly minimizes its variance, thus, pushing the probability $P_1$ to 1 and minimizing the term $ L^*$ with high probability.

$\QEDB$
\vspace*{0.25cm}

Hence, conjecture \ref{Conj: upper bound} states that if we find an estimator $\theta^*$ that reaches a sufficiently low loss value, then the respective MAP point hypothesis $W^*$ has a lower loss value with high probability. In the experimental section we provide evidence that conjecture \ref{Conj: upper bound} holds true.

\section{Related work}
There is a plethora of literature proposing different techniques and approaches for compressing the weight parameters of deep neural network \cite{DLC_survey}. Among them, some have focused on designing algorithms that sparsify the network structure \cite{Opt_brain_damage, Opt_brain_surgeon, Learning_Weight_and_Connections, dynamic_network_surgery, VD_sparsifies, l0_regularization}. Others, on reducing the cardinality of the networks parameters \cite{BinaryNet, BinnaryConnect, XNOR_net, Terniray_weights, local_reparam_trick_discrete_networks, VariationalNQ}. And some have proposed general compression frameworks with the objective to do both, sparsification and cardinality reduction \cite{deep_compression, BayesianCompression, improved_BayesianCompression}. Our entropy-constrained minimization objective \eqref{Eq: DNN min objective} (and its continuous relaxation \eqref{Eq: DNN min variational cont relax}) can be interpreted as a generalization of those works. Either maximizing sparsity or minimizing the cardinality of the weight values, both imply minimization of our derived entropy-term. However, our framework allows to also learn neural networks that are neither sparse nor their set of weight values have low cardinality, but still require low bit-lengths to represent them. Thus, our framework is able learn a larger set of networks at no cost of increase in complexity.

Other related work have also focused on designing regularizers that lower the implicit information entailed in the networks parameters, based on minimizing a variational lower bound objective \cite{Bits_back_hinton, variational_dropout, weight_info_dropout}. Although minimizing the variational lower bound is also well motivated from a MDL point of view \cite{VL_bits_back}, the resulting coding scheme is often impractical for real world scenarios. Therefore, \cite{soft_weight_sharing, BayesianCompression, improved_BayesianCompression} focused on designing suitable priors and posteriors that allow to apply practical coding schemes on the weight parameters after the variational lower bound has been minimized. This includes a final step where a lossless entropy coder is applied to the network's parameter as proposed by \cite{deep_compression}. Therefore, their proposed framework does only \textit{implicitly} minimize the resulting bit-lengths of the weight values. In contrast, our entropy-constrained objective \eqref{Eq: DNN min objective} \textit{explicitly} states the bit-size of the resulting network. Moreover, the entropy regularizer of our derived continuous relaxation \eqref{Eq: Cont relax entropy} can be rewritten as
\begin{align*}
nH(P) & = KL(q(\theta)||p(\theta)) + H(q(\theta)) \\
	& = \mathbb{E}_{q(\theta)}[-\log_2p(\theta)]
\end{align*}
with $q(\theta) = P_{\theta} =  \prod_i^n P_i$ and $p(\theta) = \prod_k^K P_k^n$, where $P_i$ and $P_k$ are defined as in \eqref{Eq: Cont relax of indicator} and \eqref{Eq: Cont relax entropy} respectively. Thus, minimizing it corresponds to minimizing the cross-entropy between the factorized posterior $q(\theta)$ and the prior $p(\theta)$ which, again, is well motivated from a practical coding point of view. In addition, the proposed prior and posterior were specifically motivated by \eqref{Eq: DNN min objective}, which led us to design non-trivial models for them. In particular, the prior $p(\theta)$ entails all weight element values in each $P_k$, and therefore implicitly takes correlations between them into account.

Finally, whilst we use similar training techniques for learning discrete neural networks as proposed by \cite{local_reparam_trick_discrete_networks, VariationalNQ}, they only focused on learning ternary and binary networks. Hence, our work can be seen as a generalization, which allows for arbitrary cardinality of the discrete set and adds an entropy term to the cost, which prioritizes non-uniform weight distributions as opposed to the training objectives stated in \cite{local_reparam_trick_discrete_networks, VariationalNQ}.

\section{Experimental results}
\subsection{Experimental setup}
In our experiments we chose to parametrize each $P_{ik}$ with the Gaussian kernel. That is,
\begin{align}
P_{ik} = & \frac{G_{\sigma_i}(w_i, \omega_k)}{Z}, \nonumber \\ 
\text{ with } G_{\sigma_i}(w_i, \omega_k) = & \frac{1}{\sigma_i \sqrt{2 \pi}} e^{-\frac{1}{2}\left(\frac{w_i - \omega_k}{\sigma_i} \right)^2} \nonumber \\ 
\text{ and } Z = & \sum_k G_{\sigma_i}(w_i, \omega_k)
\label{Eq: Gauss parametrization}
\end{align}
with parameters $\theta_i = (w_i, \omega_i, \sigma_i)$, $w_i\in \mathbb{R}, \; \omega_i \in \Omega$ and $0 < \sigma_i \in \mathbb{R}$.
We chose different set of discrete values $\Omega^l$ for each layer $l$, where we denote their respective cardinality as $|\Omega^l| = K_l \in \mathbb{N}$. Henceforth, we introduce the notation $\bar{K} = [K_{0}, K_{1}, ..., K_{l}, ..., K_{L-1}]$ in order to specify the cardinality of each layer. 
Since each layer has a different set of discrete values, we calculated the overall entropy of the network as the sum of the entropies of each individual weight element.

The experiments were performed on the MNIST dataset, using the LeNet-300-100 and LeNet-5 networks, and on the CIFAR-10 dataset, using a VGG16 network like architecture (with 15 layers), referred to as VGG-Cifar-10\footnote{\url{http://torch.ch/blog/2015/07/30/cifar.html}}. These networks are commonly used to benchmark different compression methods~\cite{deep_compression}\cite{VD_sparsifies}\cite{BayesianCompression}\cite{soft_weight_sharing}. We investigated two compression approaches:

\subsubsection{Direct entropy-constrained training} Starting from a pre-trained model, we minimized the Shannon entropy of the network parameters with the variational entropy-constrained objective \eqref{Eq: DNN min variational cont relax} (by applying algorithm \ref{Alg: noise forward pass} to each layer). We used ADAM~\cite{AdamOptimizer_Kinga2015} as the optimizer with a linearly decaying learning rate. We have also linearly increased (from 0 to up to a number less than 1) the regularization coefficient $\alpha$ in \eqref{Eq: DNN min variational cont relax}. Afterwards, in the evaluation stage, the network weights were quantized using the values of the vector $\Omega^l$, available in the $l$-th layer. We set their respective initial cardinalities $\bar{K}$ as follows: (i) LeNet-300-100, $\bar{K}=[3,3,33]$, (ii) LeNet-5, $\bar{K}=[5,5,33]$, and (iii) VGG-Cifar-10, $\bar{K}=[33,17,15,11,7,7,7,5,5,5,3,3,3,17,33]$. The quantized network performance was monitored with the classification error, compression ratio, and sparsity (calculated on the weights). 

\subsubsection{Pre-sparsifying the models} 
Firstly, we sparsified the networks by applying the sparse variational dropout technique~\cite{VD_sparsifies}. To this end, the network parameters were initialized from scratch and ADAM was applied as the optimizer with a linearly decaying learning rate. Also, we linearly increased the regularizer of the Kullback-Leibler divergence approximation during the sparsification stage (from 0 up to a number less than 1). 
After sparsification, we applied our entropy-constrained training approach to the neural networks. In this context, $\bar{K}$ was defined for each network as: (i) LeNet-300-100, $\bar{K}=[21,21,31]$, (ii) LeNet-5, $\bar{K}=[17,17,31]$, and (iii) VGG-Cifar-10, $\bar{K}=[27, .., 27, 33]$.

\textbf{Notation remark:} Henceforth, we refer to \textit{C-modelname} or \textit{Q-modelname} to either the continuous or quantized model. Furthermore \textit{ECO} refers to models that have been trained directly under the entropy-constrained objective, whereas with \textit{S+ECO} we refer to models that have been a priori sparsified. For instance, \textit{Q-LeNet-5 (S+ECO)} refers to the quantized LeNet-5 model, which has firstly been sparsified and then trained under the entropy-constrained objective.

\subsection{Verifying conjecture \ref{Conj: upper bound}}
\begin{figure}[t]
\centering
\begin{subfigure}{0.45\textwidth}
\includegraphics[width=\columnwidth,clip,keepaspectratio]{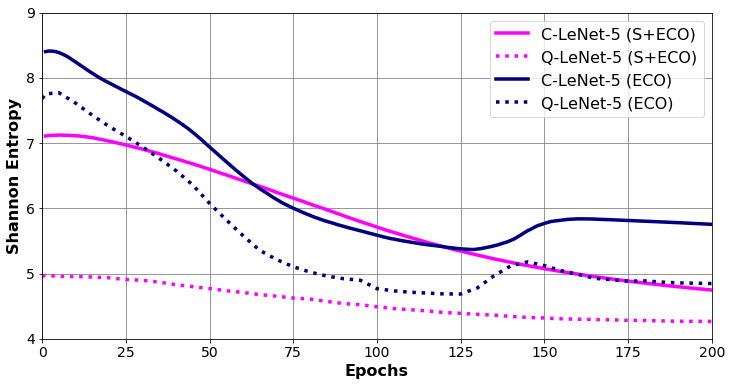} 
\caption{}
\label{Fig: MNISt H}
\end{subfigure}

\begin{subfigure}{0.45\textwidth}
\includegraphics[width=\columnwidth,clip,keepaspectratio]{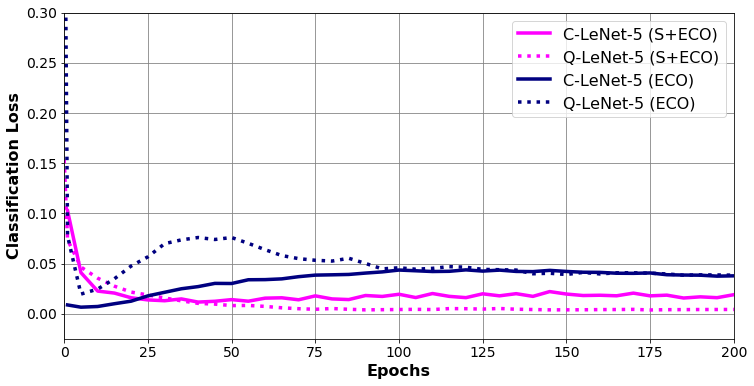} 
\caption{}
\label{Fig: MNIST train loss}
\end{subfigure}
\caption{Trend of the average log-likelihood (on the training set) and entropy of the continuous model and the quantized model during training of the LeNet-5 architecture. (a) The entropy of the quantized model is always bounded by the entropy of the continuous model during the entire training procedure. (b) The log-likelihood of the quantized model will eventually be bounded by the average likelihood of the continuous one.}
\end{figure}

\begin{figure}[t]
\includegraphics[width=\columnwidth,clip,keepaspectratio]{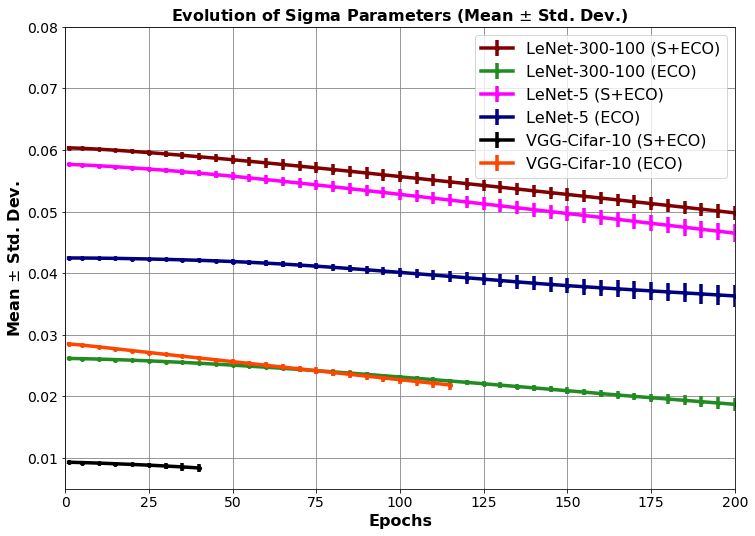} 
\caption{Trend of the variance of the continuous probability distribution during training. Each point shows the mean and standard deviation of the variance, as taken over all weight element values of the network. The plots shows how the network is trying to reduce its variance during training.}
\label{Fig: evolution sigma}
\end{figure}

Figures \ref{Fig: MNISt H}, \ref{Fig: MNIST train loss} and \ref{Fig: evolution sigma} verify some of the arguments stated in the proof idea of conjecture \ref{Conj: upper bound}. 
Firstly, figure \ref{Fig: MNISt H} shows how the entropy of the quantized model is always lower than the entropy of the continuous parametrization. Thus, the entropy of the continuous relaxation is always upper bounding the entropy of the quantized model. We also confirmed this on the other architectures. 

Secondly, from figure \ref{Fig: evolution sigma} we see how the loss automatically tends to minimize the variance of the continuous probability distribution during training. This increases the probability of selecting the quantized model (or close models) during training and, thus, minimizing its cost value as well.  

Lastly, from figure \ref{Fig: MNIST train loss} we see how indeed, this can result in the variational classification loss eventually bounding the loss of the quantized network. We verified a similar trend on the LeNet-300-100 architecture. In particular, as we predicted, by comparing figures \ref{Fig: evolution sigma} and \ref{Fig: MNIST train loss} we see how the difference between the variational and the quantized loss converge to the same value as the variance of the continuous probability distribution becomes smaller. 

\subsection{Generalizing sparsity and cardinality reduction}
\begin{figure}[t]
\includegraphics[width=\columnwidth,clip,keepaspectratio]{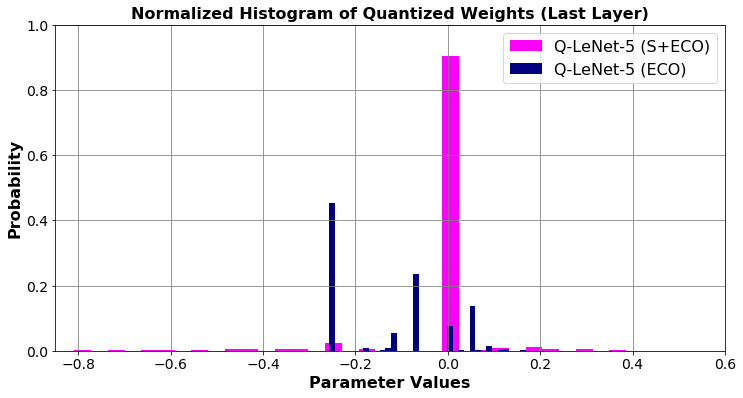} 
\caption{Probability mass distribution (PMD) of the weight element values of the last layer of the LeNet-5 model after training. The PMD of the \textit{Q-LeNet-5 (S+ECO)} is very sparse. In contrast, the PMD of the trained \textit{Q-LeNet-5 (ECO)} architecture is neither sparse nor the cardinality of its values is particularly low.}
\label{Fig: prob mass last layer}
\end{figure}
Figure \ref{Fig: prob mass last layer} shows PMDs of the last fully connected layer of the \textit{Q-LeNet-5 (ECO)} and \textit{Q-LeNet-5 (S+ECO)} models after training. Interestingly, directly minimizing the entropy-constrained objective admits solutions where the resulting weight distribution is not necessarily sparse, neither its cardinality is particularly lower than at the starting point. In contrast, by pre-sparsified the network we constrain the set of solutions to only those which are close to the spike-and-slab distribution.

Nevertheless, form table \ref{Tbl: Compression gains} we see that our entropy term is also strongly encouraging sparsity, since overall we attain similar pruning results as current state-of-the-art pruning techniques. 

\subsection{Network compression}
\begin{figure}[t]
\includegraphics[width=\columnwidth,clip,keepaspectratio]{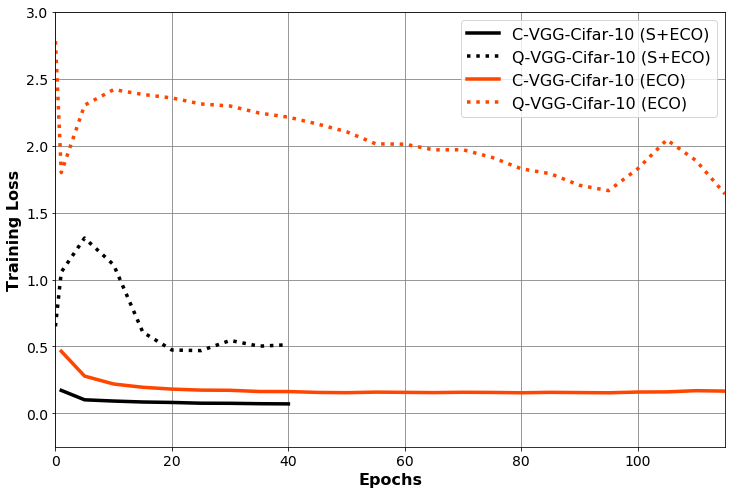} 
\caption{Trend of the classification loss of the VGG architecture when trained on the CIFAR10 data set. The plot shows how the continuous loss does not upper bound the quantized loss during training. This may explain the loss in accuracy performance in table \ref{Tbl: Compression gains} as compared to previous literature. However, we see a correlation between the variance and the difference on the loss values, in that lower variance significantly reduces the difference (see figure \ref{Fig: evolution sigma}).}
\label{Fig: CIFAR train loss}
\end{figure}

\begin{table}
\caption{Compression gains attained from training the networks under the entropy-constrained minimization objective. We also compare the results to other compression techniques.}
\centering
\begin{tabular}{|l|ccc|}
	\hline
	Model &  Error [ $\%$ ] & $\frac{|W~\neq~0|}{|~W~|}$ [ $\%$ ] &  CR  \\
	\hline	
	\multicolumn{4}{|c|}{{\bf S + ECO}}  \\
	\hline
	LeNet-300-100  			& 1.80     		& 3.28 	   		  & 92		 \\
	LeNet-5           		& 0.90          & 1.11 	          & 209		 \\
	VGG-Cifar-10 			& 10.91         & 6.77 	          & \textbf{71}		 \\	
	\hline	
	\multicolumn{4}{|c|}{{\bf ECO}}  \\
	\hline	
	LeNet-300-100 			& 2.24       & 4.03 	   	  & \textbf{102}		 \\
	LeNet-5           		& 0.97       & 1.94 	      & \textbf{235}		 \\
	VGG-Cifar-10  			& 14.27      & 6.90  	      & 95 		 \\	
	\hline
	\hline
	\multicolumn{4}{|c|}{ Deep Compression~\cite{deep_compression} }  \\
	\hline
	LeNet-300-100 			& 1.58     		  & 8  	   		  & 40		 \\
	LeNet-5           		& 0.74            & 8 	          & 39 		 \\
	\hline
	\multicolumn{4}{|c|}{ Soft Weight Sharing~\cite{soft_weight_sharing} }  \\
	\hline
	LeNet-300-100 			& 1.94     		  & 4.3  	   		  & 64		 \\
	LeNet-5           		& 0.97            & 0.5 	          & 162 		 \\
	\hline
	\multicolumn{4}{|c|}{ Sparse Variational Dropout~\cite{VD_sparsifies} }  \\
	\hline
	LeNet-300-100 			& 1.92    		& 1.47  	  & 68 		 \\
	LeNet-5           		& 0.75          & 0.35	      & 280		 \\
	VGG-Cifar-10  			& 7.30          & 1.53 	      & 65 		 \\
	\hline
	\multicolumn{4}{|c|}{ Bayesian Compression~\cite{BayesianCompression} }  \\
	\hline
	LeNet-300-100 			& 1.80     		  & 2.20 	   		  & 113 		 \\
	LeNet-5           		& 1.00            & 0.60	          & 771		 \\
	VGG-Cifar-10  			& 9.20            & 5.50 	          & 116 		 \\
	
	\hline
\end{tabular}
\label{Tbl: Compression gains}
\end{table}

Table \ref{Tbl: Compression gains} summarizes the achieved compression gains. As we can see, the results are comparable to the current state-of-the-art compression techniques. 
Interestingly, pre-sparsifying the VGG network seems to significantly improve its final accuracy. We believe that this is due to the variance of the continuous model, which may be too high during the training procedure. For large deep networks with millions of parameters such as the VGG model, the probability that the quantized model $P_1$ (or close models to it) are sampled during training reduces significantly. Consequently, the variational loss value was not upper bounding the cost of the quantized network during training, as can be seen in figure \ref{Fig: CIFAR train loss}. However, notice that the variance of the pre-sparsified VGG network is significantly lower during the entire training procedure (figure \ref{Fig: evolution sigma}), hence, closing the gap between the variational and quantized loss and consequently attaining higher accuracies on the quantized model. 

As a side note we want to stress, that the dependency between $P_1$ and the number of parameters in the network is not reflected in figure \ref{Fig: evolution sigma}. Therefore, comparisons between the variances across architectures are not informative.

\section{Conclusion}
We formalized the task of neural network compression under the Minimum Description Length (MDL) principle, which resulted in an entropy-constrained minimization objective for training deep neural networks. We argued that this objective generalizes many of the compression techniques proposed in the literature, in that it automatically encourages pruning and cardinality reduction of the weight elements. Lastly, we showed that our compression framework achieves x102, x235 and x71 compression gains on the LeNet-300-100, LeNet-5 and VGG-CIFAR-10 architecture, which is comparable to current state-of-the-art compression techniques.

In future work we will consider improvements to our method. For instance, we believe that we can attain better accuracy-compression results by adding regularizers to the loss that encourage learning network distributions with small variance (since the proper minimization of the discrete objective loss is guaranteed).
Furthermore, we will investigate the use of entropy-based compression techniques in federated learning scenarios and for improving the interpretability \cite{MonDSP18, SamITU18b} of deep models.

\section*{Acknowledgement}
This work was supported by the Fraunhofer Society through the MPI-FhG collaboration project ``Theory \& Practice for Reduced Learning Machines''. This research was also supported by the German Ministry for Education through the Berlin Big Data Center under Grant 01IS14013A and the Berlin Center for Machine Learning under Grant 01IS18037I, and by the Institute for Information \& Communications Technology Promotion and funded by the Korea government (MSIT) (No. 2017-0-01779 and No. 2017-0-00451).

\bibliographystyle{ieeetr}
\bibliography{../../References}

\begin{thebibliography}{10}

\bibitem{DeepLearning}
Y.~LeCun, Y.~Bengio, and G.~E. Hinton, ``Deep learning,'' {\em Nature},
  vol.~521, no.~7553, pp.~436--444, 2015.

\bibitem{DLC_survey}
Y.~Cheng, D.~Wang, P.~Zhou, and T.~Zhang, ``A survey of model compression and
  acceleration for deep neural networks,'' {\em arXiv:1710.09282}, 2017.

\bibitem{Random_labels}
C.~Zhang, S.~Bengio, M.~Hardt, B.~Recht, and O.~Vinyals, ``Understanding deep
  learning requires rethinking generalization,'' {\em arXiv:1611.03530}, 2016.

\bibitem{Rissanen_MDL}
J.~Rissanen, ``Paper: Modeling by shortest data description,'' {\em
  Automatica}, vol.~14, no.~5, pp.~465--471, 1978.

\bibitem{MDL}
P.~Gr{\"u}nwald and J.~Rissanen, {\em The Minimum Description Length
  Principle}.
\newblock Adaptive computation and machine learning, MIT Press, 2007.

\bibitem{federated_learning}
H.~B. McMahan, E.~Moore, D.~Ramage, and B.~A. y~Arcas, ``Federated learning of
  deep networks using model averaging,'' {\em arXiv:1602.05629}, 2016.

\bibitem{SatArXiv18}
F.~Sattler, S.~Wiedemann, K.-R. M{\"u}ller, and W.~Samek, ``Sparse binary
  compression: Towards distributed deep learning with minimal communication,''
  {\em arXiv:1805.08768}, 2018.

\bibitem{Shannon}
C.~E. Shannon, ``A mathematical theory of communication,'' {\em SIGMOBILE
  Mobile Computing and Communications Review}, vol.~5, no.~1, pp.~3--55, 2001.

\bibitem{Cover}
T.~M. Cover and J.~A. Thomas, {\em Elements of Information Theory (Wiley Series
  in Telecommunications and Signal Processing)}.
\newblock New York, NY, USA: Wiley-Interscience, 2006.

\bibitem{Wiegand_source_coding}
T.~Wiegand and H.~Schwarz, ``Source coding: Part 1 of fundamentals of source
  and video coding,'' {\em Found. Trends Signal Process.}, vol.~4, no.~1--2,
  pp.~1--222, 2011.

\bibitem{deep_compression}
S.~Han, H.~Mao, and W.~J. Dally, ``Deep compression: Compressing deep neural
  network with pruning, trained quantization and huffman coding,'' {\em
  arXiv:1510.00149}, 2015.

\bibitem{universal_succint_dnn_compression}
S.~Basu and L.~R. Varshney, ``Universal source coding of deep neural
  networks,'' in {\em Data Compression Conference (DCC)}, pp.~310--319, 2017.

\bibitem{Universal_dnn_compression}
Y.~Choi, M.~El{-}Khamy, and J.~Lee, ``Universal deep neural network
  compression,'' {\em arXiv:1802.02271}, 2018.

\bibitem{Simon_lossless_dnn_compression1}
S.~Wiedemann, K.-R. M{\"{u}}ller, and W.~Samek, ``Compact and computationally
  efficient representation of deep neural networks,'' {\em arXiv:1805.10692},
  2018.

\bibitem{MDL_coding}
A.~Barron, J.~Rissanen, and B.~Yu, ``The minimum description length principle
  in coding and modeling,'' {\em IEEE Transactions on Information Theory},
  vol.~44, no.~6, pp.~2743--2760, 2006.

\bibitem{MDL_assymptotic_conv}
Q.~Xie and A.~R. Barron, ``Asymptotic minimax regret for data compression,
  gambling, and prediction,'' {\em IEEE Transactions on Information Theory},
  vol.~46, no.~2, pp.~431--445, 2000.

\bibitem{DL_tricks_trade}
Y.~LeCun, L.~Bottou, G.~B. Orr, and K.-R. M{\"{u}}ller, ``Efficient backprop,''
  in {\em Neural Networks: Tricks of the Trade - Second Edition, Springer LNCS
  7700}, pp.~9--48, Springer, 2012.

\bibitem{variational_dropout}
D.~P. Kingma, T.~Salimans, and M.~Welling, ``Variational dropout and the local
  reparameterization trick,'' in {\em Advances in Neural Information Processing
  Systems (NIPS)}, pp.~2575--2583, 2015.

\bibitem{weight_info_dropout}
A.~Achille and S.~Soatto, ``On the emergence of invariance and disentangling in
  deep representations,'' {\em arXiv:1706.01350}, 2017.

\bibitem{VD_sparsifies}
D.~Molchanov, A.~Ashukha, and D.~Vetrov, ``Variational dropout sparsifies deep
  neural networks,'' in {\em International Conference on Machine Learning
  (ICML)}, pp.~2498--2507, 2017.

\bibitem{BayesianCompression}
C.~{Louizos}, K.~{Ullrich}, and M.~{Welling}, ``{Bayesian Compression for Deep
  Learning},'' in {\em Advances in Neural Information Processing Systems
  (NIPS)}, pp.~3290--3300, 2017.

\bibitem{Opt_brain_damage}
Y.~L. Cun, J.~S. Denker, and S.~A. Solla, ``Optimal brain damage,'' in {\em
  Advances in Neural Information Processing Systems (NIPS)}, pp.~598--605,
  1990.

\bibitem{Opt_brain_surgeon}
B.~Hassibi, D.~G. Stork, and G.~J. Wolff, ``Optimal brain surgeon and general
  network pruning,'' in {\em IEEE International Conference on Neural Networks},
  pp.~293--299, 1993.

\bibitem{Learning_Weight_and_Connections}
S.~Han, J.~Pool, J.~Tran, and W.~J. Dally, ``Learning both weights and
  connections for efficient neural networks,'' in {\em Advances in Neural
  Information Processing Systems (NIPS)}, pp.~1135--1143, 2015.

\bibitem{dynamic_network_surgery}
Y.~Guo, A.~Yao, and Y.~Chen, ``Dynamic network surgery for efficient dnns,''
  {\em arXiv:1608.04493}, 2016.

\bibitem{l0_regularization}
C.~{Louizos}, M.~{Welling}, and D.~P. {Kingma}, ``{Learning Sparse Neural
  Networks through $L\_0$ Regularization},'' {\em arXiv:1712.01312}, 2017.

\bibitem{BinaryNet}
M.~Courbariaux and Y.~Bengio, ``Binarynet: Training deep neural networks with
  weights and activations constrained to +1 or -1,'' {\em arXiv:1602.02830},
  2016.

\bibitem{BinnaryConnect}
M.~Courbariaux, Y.~Bengio, and J.~David, ``Binaryconnect: Training deep neural
  networks with binary weights during propagations,'' {\em arXiv:1511.00363},
  2015.

\bibitem{XNOR_net}
M.~Rastegari, V.~Ordonez, J.~Redmon, and A.~Farhadi, ``Xnor-net: Imagenet
  classification using binary convolutional neural networks,'' {\em
  arXiv:1603.05279}, 2016.

\bibitem{Terniray_weights}
F.~Li, B.~Zhang, and B.~Liu, ``Ternary weight networks,'' {\em
  arXiv:1605.04711}, 2016.

\bibitem{local_reparam_trick_discrete_networks}
O.~Shayer, D.~Levi, and E.~Fetaya, ``Learning discrete weights using the local
  reparameterization trick,'' {\em arXiv:1710.07739}, 2017.

\bibitem{VariationalNQ}
J.~Achterhold, J.~M. K{\"o}hler, A.~Schmeink, and T.~Genewein, ``Variational
  network quantization,'' in {\em International Conference on Representation
  Learning (ICLR)}, 2018.

\bibitem{improved_BayesianCompression}
M.~{Federici}, K.~{Ullrich}, and M.~{Welling}, ``{Improved Bayesian
  Compression},'' {\em arXiv:1711.06494}, 2017.

\bibitem{Bits_back_hinton}
G.~E. Hinton and D.~van Camp, ``Keeping the neural networks simple by
  minimizing the description length of the weights,'' in {\em Conference on
  Computational Learning Theory (COLT)}, pp.~5--13, 1993.

\bibitem{VL_bits_back}
H.~V. Antti~Honkela, ``Variational learning and bits-back coding: An
  information-theoretic view to bayesian learning,'' 2004.

\bibitem{soft_weight_sharing}
K.~{Ullrich}, E.~{Meeds}, and M.~{Welling}, ``{Soft Weight-Sharing for Neural
  Network Compression},'' {\em arXiv:1702.04008}, 2017.

\bibitem{AdamOptimizer_Kinga2015}
D.~Kinga and J.~B. Adam, ``A method for stochastic optimization,'' in {\em
  International Conference on Learning Representations (ICLR)}, 2015.

\bibitem{MonDSP18}
G.~Montavon, W.~Samek, and K.-R. M{\"u}ller, ``Methods for interpreting and
  understanding deep neural networks,'' {\em Digital Signal Processing},
  vol.~73, pp.~1--15, 2018.

\bibitem{SamITU18b}
W.~Samek, T.~Wiegand, and K.-R. M{\"u}ller, ``Explainable artificial
  intelligence: Understanding, visualizing and interpreting deep learning
  models,'' {\em ITU Journal: ICT Discoveries - Special Issue 1 - The Impact of
  Artificial Intelligence (AI) on Communication Networks and Services}, vol.~1,
  no.~1, pp.~39--48, 2018.

\end{thebibliography}

\end{document}